\documentclass{article} 
\usepackage[final]{colm2026_conference}

\usepackage{microtype}
\usepackage{hyperref}
\usepackage{url}
\usepackage{booktabs}
\usepackage{lineno}
\usepackage{amsmath,amssymb,amsthm}
\usepackage{multirow}
\usepackage{graphicx}
\usepackage{subcaption}
\usepackage[ruled,vlined,linesnumbered]{algorithm2e}
\usepackage[utf8]{inputenc}
\usepackage{xcolor}
\usepackage[most]{tcolorbox}
\usepackage{caption}
\usepackage{xspace}

\newtheorem{remark}{Remark}

\definecolor{darkblue}{rgb}{0, 0, 0.5}
\hypersetup{colorlinks=true, citecolor=darkblue, linkcolor=darkblue, urlcolor=darkblue}

\newcommand{\rfmt}{r_{\mathrm{fmt}}}

\newcommand{\cnt}{\mathrm{cnt}}
\newcommand{\Smax}{S_{\max}}

\title{CuSearch: Curriculum Rollout Sampling via Search Depth for Agentic RAG}

\author{%
  Jianghan Shen$^{1,2}$,\;
  Siqi Luo$^{2}$,\;
  Xinyu Cheng$^{1,2}$,\;
  Jing Xiong$^{4}$,\;
  Yue Li$^{3}$,\;
  Jiyao Liu$^{2}$ \\[2pt]
  \textbf{Jiashi Lin$^{2}$,\;
  Yirong Chen$^{2}$,\;
  Junjun He$^{2,5}$\thanks{Corresponding author: \texttt{hejunjun@pjlab.org.cn}}} \\[4pt]
  \normalfont $^1$Nanjing University \quad
  $^2$Shanghai Artificial Intelligence Laboratory \\[2pt]
  $^3$Peking University \quad
  $^4$University of Hong Kong \quad
  $^5$Shanghai Innovation Institute \quad
}

\begin{document}

\ifcolmsubmission
\linenumbers
\fi

\maketitle

% ── Abstract ─────────────────────────────────────────────────────────────────
\begin{abstract}
Reinforcement Learning with Verifiable Rewards (\textbf{RLVR}) has emerged as a promising paradigm for training agentic retrieval-augmented generation (\textbf{RAG}) systems from outcome-only supervision.
Most existing methods optimize policies from uniformly sampled rollouts, implicitly treating all trajectories as equally informative.
However, trajectories differ substantially in search depth and are therefore not equally informative: deeper-search trajectories contain more retrieval decision points and provide denser direct supervision for the retrieval sub-policy.
Moreover, this heterogeneity grows over training as the within-batch depth distribution shifts toward higher values, yet uniform rollout sampling remains blind to this shift.
To address this, we propose \textbf{CuSearch}, a curriculum rollout sampling framework built on \textbf{Search-Depth Greedy Allocation (SDGA)}, a batch-level operator that reallocates a fixed update budget toward deeper-search trajectories.
\textbf{SDGA-Auto} always targets the deepest available trajectories in the current batch, yielding an implicit training-aligned curriculum as the depth distribution shifts upward.
\textbf{SDGA-Phase} explicitly advances the curriculum threshold as deeper trajectories become sufficiently abundant.
Experiments across model types and retrieval frameworks show that \textbf{CuSearch} consistently improves performance, achieving up to \textbf{11.8} exact-match points over standard \textbf{GRPO} on \textbf{ZeroSearch}.
These results establish per-trajectory search depth as a reliable, annotation-free proxy for retrieval supervision density in RLVR-based agentic RAG training.
The code is available at \url{https://github.com/MrToser/CuSearch}.
\end{abstract}

% ── Main Sections ─────────────────────────────────────────────────────────────
\section{Introduction}

Reinforcement Learning with Verifiable Rewards (RLVR) has become an effective approach for training large language models (LLMs) from outcome-only supervision~\citep{ouyang2022training,shao2024grpo,guo2025deepseekr1}.
This paradigm has been extended to agentic retrieval-augmented generation (RAG)~\citep{lewis2020rag}, where models learn to decide when to retrieve and what to query during reasoning.
For knowledge-intensive tasks, model parameters alone cannot cover all relevant facts, making the ability to actively retrieve external information at inference time essential.
Earlier systems relied on heuristic triggers~\citep{jiang2023flare,trivedi2023ircot} or supervised fine-tuning on expert trajectories~\citep{asai2024selfrag,nakano2021webgpt}. Contemporary RLVR-based methods~\citep{jin2025searchr1,shi2025r1searcher,chen2025research,shen2025zerosearch,qwen2025tongyideepresearch} now optimize retrieval policies solely through final answer rewards, bypassing manual design or expert traces.

Despite this potential, existing methods optimize policies from uniformly sampled rollouts~\citep{shao2024grpo,schulman2017ppo}, implicitly treating all trajectories as equally informative.
For the same prompt, trajectories can vary substantially in search depth, defined as the number of valid search steps they issue.
Search depth provides a simple structural indicator of how many retrieval decision points a trajectory exposes and thus serves as a practical proxy for how much direct supervision it provides for the retrieval sub-policy.
Treating these unequally informative trajectories as equivalent allocates the update budget indiscriminately. This dilutes the influence of deeper-search trajectories and assigns optimization budget to rollouts with minimal retrieval learning value.

Existing work on non-uniform sample usage has largely operated at the prompt level, filtering or reallocating the training budget based on prompt-level signals such as reward variance, estimated difficulty, or expected policy improvement~\citep{yu2025dapo,zhang2025ar3po,wang2025heapa,li2026actorcurator,chen2025agentfrontier}.
By aggregating statistics across rollouts for the same prompt, these approaches discard within-batch variation in individual trajectory search behavior.
Process-based supervision~\citep{lightman2023lets,xiong2025raggym} provides denser intermediate feedback but requires additional annotation or auxiliary models.
None of these approaches exploit search depth, an annotation-free and trajectory-intrinsic structural signal for retrieval supervision density.

\begin{figure}[t]
\centering
% \vspace{-1mm}
\includegraphics[width=\linewidth]{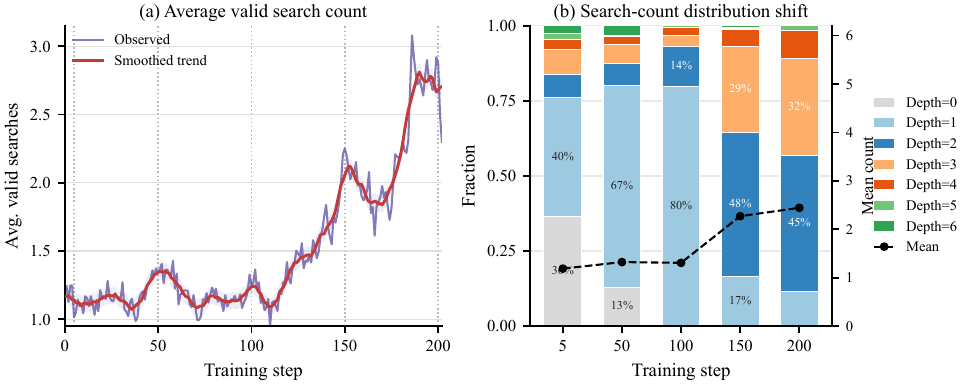}
% \vspace{-4mm}
\caption{(a) Average search count per trajectory increases over training under answer-correctness reward alone. (b) The within-batch depth distribution shifts toward higher values as training advances.}
\label{fig:search-dist}
% \vspace{-6mm}
\end{figure}

Over the course of training, answer-correctness reward reinforces retrieval behavior, and per-trajectory search depth increases monotonically.
As shown in Figure~\ref{fig:search-dist}, the within-batch depth distribution shifts toward larger values as RLVR progresses, deepening the informativeness gap across trajectories and making the misallocation of the update budget increasingly consequential.
This temporal shift motivates a trajectory-level curriculum that concentrates the update budget on the deepest available trajectories within each batch and naturally advances as the depth distribution shifts upward.

% \vspace{-4mm}

Motivated by these observations, we propose \textbf{CuSearch}, a curriculum rollout sampling framework for agentic RAG.
At the core of CuSearch is \textbf{Search-Depth Greedy Allocation (SDGA)}, a batch-level operator that selects $K$ trajectories from a larger pool of $N \cdot G$ rollouts, restricting reward normalization, advantage estimation, and gradient updates to this subset.
Crucially, SDGA selects by search depth rather than by reward.
It prefers trajectories that expose more retrieval decision points, even when final rewards are similar.
We instantiate SDGA in two variants.
\textbf{SDGA-Auto} always targets the deepest available trajectories in the current batch, yielding an implicit training-aligned curriculum as the depth distribution shifts upward.
\textbf{SDGA-Phase} makes this curriculum explicit, advancing a phase threshold irreversibly toward deeper trajectories as they become sufficiently abundant in the rollout pool.

Experiments across model types and retrieval frameworks show that CuSearch consistently improves final performance under comparable rollout budgets.
Gains reach up to \textbf{11.8} exact-match points over standard GRPO on ZeroSearch, with substantial margins over prompt-level adaptive methods such as DAPO and AR3PO.

% \vspace{4mm}
\textbf{Our contributions are:}
\begin{enumerate}
    \item We identify and empirically characterize a training dynamic in agentic RAG: under answer-correctness reward alone, per-trajectory search depth increases over training, increasing the heterogeneity in retrieval supervision density across trajectories.
    \item We propose \textbf{CuSearch}, a framework built on the \textbf{SDGA} operator, with two instantiations. \textbf{SDGA-Auto} is a stateless greedy variant that always targets the deepest available trajectories, yielding an implicit training-aligned curriculum as the depth distribution shifts upward. \textbf{SDGA-Phase} is an explicit phase-adaptive variant that advances the curriculum threshold as deeper trajectories become sufficiently abundant. Both reallocate a fixed update budget toward deeper-search trajectories without additional annotation or auxiliary models.
    \item We demonstrate consistent gains across two model families, two retrieval frameworks, and seven benchmarks. We further identify two factors governing gain magnitude: search depth growth during training as the primary driver, and retrieval-environment stochasticity as a secondary amplifier.
\end{enumerate}

\section{Related Work}

% \vspace{-4pt}
\subsection{Agentic RAG Training with Reinforcement Learning}

Early RAG methods retrieve a fixed set of documents and condition generation on them statically \citep{lewis2020rag}. 
Subsequent agentic retrieval methods interleave reasoning and retrieval over multiple steps, enabling models to gather evidence adaptively during inference \citep{singh2025agentragsurvey, wei2022cot}. 
Representative examples include ReAct \citep{yao2023react}, FLARE \citep{jiang2023flare}, and IRCoT \citep{trivedi2023ircot}. 
While these approaches improve retrieval--reasoning coordination, their interaction protocols are typically predefined or rely on heuristic retrieval control, rather than being optimized end-to-end from answer rewards.

Recent work extends RLVR to agentic RAG and directly optimizes retrieval behavior from outcome-only supervision. 
Search-R1 \citep{jin2025searchr1}, R1-Searcher \citep{shi2025r1searcher}, ZeroSearch \citep{shen2025zerosearch}, ReZero \citep{dao2025rezero}, RAG-Gym \citep{xiong2025raggym}, and ReasonRAG \citep{xu2025reasonrag} study RL-based retrieval optimization through different training recipes, retrieval environments, and reward formulations. 
Our work is complementary to this line of research. Rather than introducing a new retrieval environment or auxiliary reward signal, we study how a fixed update budget should be allocated across trajectories during training. 
Specifically, we use per-trajectory search depth as an annotation-free structural proxy for trajectory informativeness, and prioritize trajectories that expose denser supervision signal for the retrieval sub-policy.

% \vspace{-4pt}
\subsection{Curriculum and Adaptive Sampling in RL}
\label{sec:related-adaptive}

Curriculum learning improves optimization by prioritizing training samples with greater learning value \citep{bengio2009curriculum}.
Prioritized experience replay \citep{schaul2015prioritized} selects transitions by temporal-difference error, and DART-Math \citep{tong2024dartmath} applies difficulty-aware rejection sampling in supervised fine-tuning.
Recent RLVR methods extend this principle to prompt-level allocation \citep{yu2025dapo, zhang2025ar3po, wang2025heapa, li2026actorcurator, chen2025agentfrontier}, filtering or reallocating gradient budget based on prompt-level signals such as reward variance, estimated difficulty, or expected policy improvement \citep{liu2025exploredataleft, wang2025nopromptleft, bai2025odf}.
However, these approaches primarily focus on prompt-level allocation, loss-based prioritization, or offline data construction. None of them exploit the per-trajectory structural signal that governs retrieval informativeness within a training batch.
Complementary work studies rollout structure: Depth-Breadth Synergy \citep{sun2025depthbreadth} balances reasoning depth and diversity, InfoFlow \citep{chen2025infoflow} optimizes reward density, and process supervision \citep{lightman2023lets, xiong2025raggym} provides step-level annotations.

CuSearch operates at the trajectory level rather than the prompt level, and uses search depth as a structural signal rather than a reward-derived one.
Prompt-level methods decide which prompts enter training, while CuSearch decides, from the rollouts already generated, which trajectories most directly cover the retrieval decision space.
Trajectory-level supervision has also been explored for on-policy distillation, replacing token-level probability matching with verbal-score matching~\citep{xiong2026ovd}.
These two operations are orthogonal, and the gains of CuSearch over prompt-level baselines under matched compute confirm that trajectory-level structural selection addresses a distinct dimension of the training allocation problem.
% \vspace{-8pt}
\section{Method}

\begin{figure}[t]
\centering
\includegraphics[width=\linewidth]{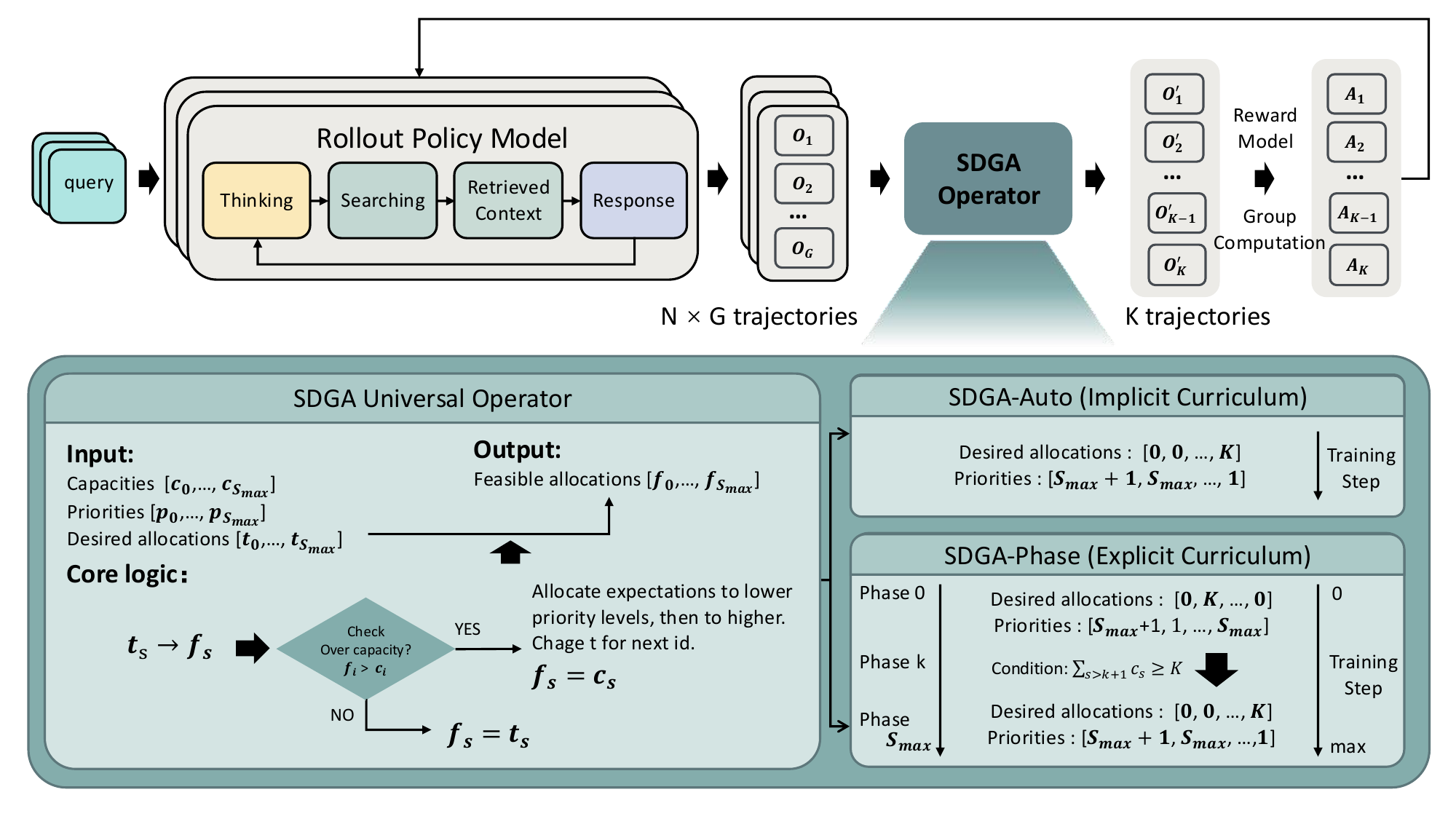}
% \fbox{\parbox[c][5cm][c]{\linewidth}{\centering\small\textit{[method\_overview.pdf]}}}
\caption{Overview of CuSearch.
\textit{Top}: A query generates $N \cdot G$ rollouts. SDGA then selects $K$ by search depth for gradient updates.
\textit{Bottom left}: SDGA greedily allocates budget $K$ across depth buckets by priority and capacity.
\textit{Bottom right}: SDGA-Auto always targets the deepest available bucket, with selections advancing deeper as the depth distribution shifts upward during training. SDGA-Phase explicitly advances the target bucket as deeper trajectories become sufficiently abundant.}
\label{fig:method-overview}
% \vspace{-4mm}
\end{figure}

% \vspace{-4pt}
\subsection{Problem Formulation}

\textbf{GRPO training framework.}
CuSearch is built on top of GRPO~\citep{shao2024grpo}.
At each training step, given a batch of $N$ prompts $\{x_i\}_{i=1}^N$, GRPO samples $G$ rollouts $\{\tau_{i,g}\}_{g=1}^G$ for each prompt.
It then computes within-group advantages $A_{i,g}$ from relative rewards and optimizes
% \vspace{-4pt}
\begin{equation}
  \mathcal{L}_{\mathrm{GRPO}}(\theta)
  =
  -\frac{1}{N \cdot G}\sum_{i=1}^{N}\sum_{g=1}^{G}
  \left[
    \mathcal{J}(\rho_{i,g}, A_{i,g})
    - \beta \,\mathrm{KL}_{i,g}\!\left(\pi_\theta \,\|\, \pi_{\mathrm{ref}}\right)
  \right],
  \label{eq:grpo-standard}
\end{equation}
% \vspace{-4pt}
where $\mathcal{J}(\rho, A) = \min\!\left(\rho A,\;\mathrm{clip}(\rho,\,1{-}\varepsilon,\,1{+}\varepsilon)A\right)$ is the clipped surrogate objective, $\rho_{i,g} = \pi_\theta(\tau_{i,g}\mid x_i)/\pi_{\mathrm{old}}(\tau_{i,g}\mid x_i)$ is the likelihood ratio, and $\varepsilon$, $\beta$ are the clipping and KL penalty coefficients, respectively. Standard GRPO uses all $N \cdot G$ rollouts uniformly.
% \vspace{-40pt}

\textbf{CuSearch modification.}
Let
\[
\mathcal{T}=\{\tau_{i,g}\}_{i=1,g=1}^{N,G},
\qquad |\mathcal{T}| = N \cdot G,
\]
denote the complete rollout set sampled at one training step.
CuSearch treats each trajectory $\tau \in \mathcal{T}$ as an atomic training unit and selects a subset
\[
T \subseteq \mathcal{T},
\qquad |T| = K \le N \cdot G,
\]
using the Search-Depth Greedy Allocation module introduced in Section~\ref{sec:sdga} and illustrated in Figure~\ref{fig:method-overview}.
The policy objective is then evaluated only on the selected subset:
\begin{equation}
  \mathcal{L}_{\mathrm{CuSearch}}(\theta)
  =
  -\frac{1}{K}\sum_{\tau \in T}
  \left[
    \mathcal{J}(\rho_\tau, A_\tau)
    - \beta \,\mathrm{KL}_\tau\!\left(\pi_\theta \,\|\, \pi_{\mathrm{ref}}\right)
  \right],
  \label{eq:grpo-cusearch}
\end{equation}
where $\rho_\tau$, $A_\tau$, and $\mathrm{KL}_\tau(\pi_\theta \| \pi_{\mathrm{ref}})$ denote the likelihood ratio, advantage, and KL penalty associated with trajectory $\tau$, respectively.
Advantages are computed from trajectories in $T$ via within-group reward normalization, and groups with no retained trajectory are excluded from the update.
CuSearch thus changes which rollouts participate in reward normalization, advantage estimation, and gradient updates, without altering the GRPO optimization form.

\textbf{Search-count-based selection problem.}
Each trajectory $\tau$ has a search count $\mathrm{cnt}(\tau) \in \{0, \ldots, S_{\max}\}$, and the goal is to construct $T \subseteq \mathcal{T}$ with $|T|=K$ by prioritizing trajectories with higher search depth, as these contain more retrieval decision points and provide denser direct gradient coverage for the retrieval sub-policy.
Appendix~\ref{app:theory} elaborates on this connection.

% \vspace{-4pt}
\subsection{Search-Depth Greedy Allocation (SDGA)}
% \vspace{-2pt}
\label{sec:sdga}

\textbf{Bucketization by search depth.}
We first partition the rollout set $\mathcal{T}$ into search-depth buckets.
For each $s \in \{0,\ldots,S_{\max}\}$, let
\[
\mathcal{B}_s = \{\tau \in \mathcal{T} \mid \mathrm{cnt}(\tau)=s\},
\qquad
c_s = |\mathcal{B}_s|,
\]
where $c_s$ is the capacity of bucket $s$ and $\mathrm{cnt}(\tau)$ counts the well-formed \texttt{<search>}$\ldots$\texttt{</search>} sequences with non-empty retrieved content in trajectory $\tau$.
A rollout allocation policy specifies how many trajectories should be retained from each bucket under the global budget constraint $\sum_{s=0}^{S_{\max}} f_s = K$.

\textbf{Greedy allocation by priority.}
Given bucket capacities $\{c_s\}_{s=0}^{S_{\max}}$ and a priority ordering $\{p_s\}_{s=0}^{S_{\max}}$ over buckets, SDGA allocates the update budget $K$ greedily according to priority, producing a feasible allocation $\{f_s\}_{s=0}^{S_{\max}}$ such that
\[
0 \le f_s \le c_s,
\qquad
\sum_{s=0}^{S_{\max}} f_s = K.
\]
The algorithm proceeds in priority order: for each bucket $s$, it attempts to allocate as many trajectories as the bucket can supply.
If the desired allocation exceeds the bucket capacity $c_s$, the shortfall is redistributed to remaining buckets in priority order.
This greedy strategy ensures that higher-priority buckets are filled first, while respecting capacity constraints.
The priority ordering determines the curriculum: different instantiations of SDGA correspond to different choices of $\{p_s\}$.
Additional implementation details are provided in Appendix~\ref{app:sdga}.

\begin{algorithm}[H]
\SetAlgoLined
\caption{Search-Depth Greedy Allocation (SDGA)}
\label{alg:sdga}
\KwIn{Capacities $\{c_s\}_{s=0}^{S_{\max}}$, desired allocations $\{t_s\}_{s=0}^{S_{\max}}$ with $\sum_s t_s = K$, priorities $\{p_s\}_{s=0}^{S_{\max}}$}
\KwOut{Feasible allocations $\{f_s\}_{s=0}^{S_{\max}}$}

$f_s \leftarrow t_s$ for all $s$\;
$\sigma \leftarrow \mathrm{argsort}(\{p_s\})$\;

\ForEach{$s$ in order of $\sigma$}{
    \If{$f_s > c_s$}{
        $d \leftarrow f_s - c_s$;\quad $f_s \leftarrow c_s$\;
        \tcp{redistribute $d$: lower-priority buckets first, then higher-priority in reverse}
        \ForEach{$r \ne s$ (lower-priority order, then higher-priority in reverse)}{
            \If{$d = 0$}{\textbf{break}\;}
            $\delta \leftarrow \min(c_r - f_r,\;d)$;\quad $f_r \leftarrow f_r + \delta$;\quad $d \leftarrow d - \delta$\;
        }
    }
}
\Return $\{f_s\}_{s=0}^{S_{\max}}$\;
\end{algorithm}

\textbf{Trajectory instantiation.}
After obtaining feasible bucket allocations $\{f_s\}$, CuSearch samples exactly $f_s$ trajectories from each bucket $\mathcal{B}_s$ to form the selected set $T$.
Thus, SDGA operates at the bucket level, while the final rollout subset is instantiated at the trajectory level.
This design makes the allocation step lightweight and independent of the downstream GRPO optimizer.

\textbf{SDGA-Auto.}
SDGA-Auto is the stateless instantiation of SDGA.
It sets $p_s = S_{\max} - s + 1$ to give higher priority to deeper search buckets and sets $t_{S_{\max}} = K$ with $t_s = 0$ for $s < S_{\max}$ to ideally allocate all budget to the deepest bucket.
At every step, it greedily concentrates the full update budget on the deepest-search trajectories available in the current batch.
When the deepest bucket is under-capacity, the shortfall is redistributed to shallower buckets in descending order of search depth.
We deliberately decouple the selection criterion from reward to isolate the retrieval supervision signal. Reward information is incorporated downstream through within-group advantage normalization.
Although stateless, SDGA-Auto implicitly follows a curriculum aligned with the model's training dynamics: because the within-batch depth distribution shifts monotonically toward higher search counts during training, the depth level that SDGA-Auto actually concentrates on rises progressively, advancing from moderately deep trajectories early in training to the genuinely deepest trajectories as they become prevalent.
This implicit curriculum emerges naturally from the interaction between the greedy selection rule and the evolving rollout distribution, without any explicit phase management.
See Appendix~\ref{app:sdga-auto} for details.

\textbf{SDGA-Phase.}
SDGA-Phase is the phase-adaptive instantiation of SDGA, motivated by curriculum learning~\citep{bengio2009curriculum}.
It maintains a monotonically advancing phase threshold $k \in \{0, \ldots, S_{\max}-1\}$ and sets $t_{k+1} = K$ and $t_s = 0$ for $s \ne k+1$, targeting bucket $k+1$ at phase $k$.
The priority ordering favors bucket $k+1$ first, then deeper buckets, then shallower ones.
The phase advances from $k$ to $k+1$ when $\sum_{s > k+1} c_s \ge K$, ensuring the target bucket is deep enough to provide informative retrieval supervision yet abundant enough to avoid falling back to shallow trajectories.
This design prevents the fallback to shallow-bucket selection that SDGA-Auto can suffer early in training, when the deepest bucket is sparsely populated. See Appendix~\ref{app:sdga-phase} for details.

% \vspace{-4pt}
\subsection{Implementation and Training Setup}
% \vspace{-2pt}

CuSearch is inserted as a lightweight trajectory selection layer between rollout generation and policy optimization, requiring no changes to the reward function, model architecture, or agent environment.
The complete training loop is in Appendix~\ref{app:cusearch-loop}.
We follow the standard setup from prior RLVR-based agentic RAG work, keeping the prompt template, output format, and reward design fixed so that all experiments vary only the rollout allocation strategy.
Complete training details are in Appendix~\ref{app:training-details}.

\section{Experiments}

% \vspace{-4pt}
\subsection{Experimental Setup}

% \vspace{1mm}
\textbf{Models and training.}
We use \textbf{Qwen2.5-3B}~\citep{qwen2024qwen25} as the primary backbone for ablation and analysis, \textbf{LLaMA-3.2-3B-Base}~\citep{llama2024llama3} to test cross-model generalization, and \textbf{Qwen2.5-7B}~\citep{qwen2024qwen25} to examine scalability across model sizes.
All models are trained with VERL and GRPO.
Unless otherwise specified, we set $N = 64$, $G = 3$, $K = 96$, $M = 32$, $\Smax = 5$, learning rate $= 1 \times 10^{-6}$, and $\beta = 0.001$.
All experiments are conducted on $8 \times$ A100 80G GPUs.
Unless otherwise noted, all figures in this paper report results on Qwen2.5-3B with ZeroSearch.

% \vspace{1mm}
\textbf{Retrieval frameworks.}
We evaluate CuSearch under two representative agentic retrieval settings.
\textbf{ZeroSearch}~\citep{shen2025zerosearch} uses LLM-simulated retrieval, and we adopt Alibaba-NLP/Simulation-LLM-google-7B-V2 as the retriever simulator in the main experiments.
To test whether the learned policy transfers beyond the simulator, we further evaluate it with a real search API, as detailed in Appendix~\ref{app:ablation-serper}.
\textbf{Search-R1}~\citep{jin2025searchr1} uses a chunk-based offline retrieval corpus for both training and evaluation.
Together, these two settings allow us to study CuSearch under retrieval environments with different levels of stochasticity.

% \vspace{1mm}
\textbf{Datasets and metric.}
We evaluate on seven open-domain QA benchmarks.
The single-hop datasets are NQ~\citep{kwiatkowski2019nq}, TriviaQA~\citep{joshi2017triviaqa}, and PopQA~\citep{mallen2023popqa}.
The multi-hop datasets are HotpotQA~\citep{yang2018hotpotqa}, 2WikiMultiHopQA~\citep{ho2020twowiki}, MuSiQue~\citep{trivedi2022musique}, and Bamboogle~\citep{press2023bamboogle}.
We use \textbf{Exact Match (EM)} as the evaluation metric.

% \vspace{1mm}
\textbf{Baselines.}
We compare CuSearch against both rollout-selection baselines and representative prompt-level adaptive methods.
\textbf{GRPO-Full} uses all $N \cdot G$ rollouts without trajectory selection, corresponding to the standard training protocol of ZeroSearch or Search-R1.  
\textbf{GRPO-Half} uses $K/G$ prompts with $G$ rollouts each, yielding the same $K$ trajectory updates as CuSearch but without depth-aware selection. Unlike CuSearch, it generates only $K$ rollouts directly rather than selecting from a larger pool.
\textbf{SDGA-Auto} and \textbf{SDGA-Phase} are our two search-depth-based allocation strategies, both using $\Smax = 5$ in the main experiments.
\textbf{SDGA-Anti} reverses the search-depth priority order to deliberately select shallow-search trajectories first, serving as an anti-curriculum directional ablation.
\textbf{Random} draws $K$ trajectories uniformly at random from $\mathcal{T}$, matching CuSearch in both rollout generation and update cost, so any performance gap isolates the contribution of the depth-aware selection signal.
\textbf{Top-K Reward} selects the $K$ highest-reward trajectories from $\mathcal{T}$, matching CuSearch in compute, and tests whether reward magnitude alone provides an effective selection criterion.
To position CuSearch relative to prior adaptive RL methods, we additionally compare with \textbf{DAPO}~\citep{yu2025dapo}, which removes prompts whose rollouts all receive the same reward, and \textbf{AR3PO}~\citep{zhang2025ar3po}, which adaptively reallocates rollout budget across prompts based on estimated difficulty.
Both are implemented within the same VERL/GRPO pipeline under matched total compute.

% \vspace{-4pt}
\subsection{Main Results}

\begin{table}[ht]
\caption{Main results under ZeroSearch (simulator evaluation) and Search-R1 frameworks (EM, \%).
GRPO-Full and GRPO-Half correspond to the standard training protocol of each framework without trajectory selection.
All SDGA variants use $\Smax = 5$.
\textbf{Bold} denotes the best result per model group.}
\label{tab:main}
\centering
\small
\resizebox{\linewidth}{!}{%
\begin{tabular}{llccccccccc}
\toprule
Framework & Model & Method & NQ & TQA & PopQA & HotpotQA & 2Wiki & MuSiQue & BG & Avg. \\
\midrule
\multirow{15}{*}{ZeroSearch}
& \multirow{5}{*}{LLaMA-3.2-3B-Base}
    & GRPO-Full & 37.9 & 52.8 & 46.5 & 33.2 & 32.6 & 15.8 & 16.4 & 33.6 \\
  & & GRPO-Half & 34.6 & 48.9 & 42.8 & 28.9 & 28.7 & 10.5 & 13.6 & 29.7 \\
  & & SDGA-Anti & 19.8 & 24.9 & 17.9 & 15.8 & 16.4 & 9.8 & 8.5 & 16.2 \\
  & & SDGA-Auto & 42.5 & \textbf{57.4} & 52.3 & \textbf{38.2} & \textbf{37.1} & 20.1 & \textbf{22.1} & 38.5 \\
  & & \textbf{SDGA-Phase} & \textbf{43.2} & 57.1 & \textbf{53.1} & 37.9 & 36.8 & \textbf{21.4} & 21.8 & \textbf{38.8} \\
\cmidrule{2-11}
  & \multirow{5}{*}{Qwen2.5-3B}
    & GRPO-Full & 44.3 & 60.7 & 48.1 & 44.5 & 41.2 & 32.7 & 39.8 & 44.5 \\
  & & GRPO-Half & 44.1 & 58.5 & 52.3 & 36.9 & 36.6 & 22.3 & 24.1 & 39.3 \\
  & & SDGA-Anti & 20.2 & 35.3 & 17.7 & 19.9 & 24.7 & 4.3 & 20.9 & 20.4 \\
  & & SDGA-Auto & 53.4 & 67.3 & \textbf{68.0} & 51.5 & \textbf{51.8} & 39.4 & 39.1 & 52.9 \\
  & & \textbf{SDGA-Phase} & \textbf{55.1} & \textbf{72.4} & 67.9 & \textbf{52.8} & 49.5 & \textbf{44.9} & \textbf{51.4} & \textbf{56.3} \\
\cmidrule{2-11}
  & \multirow{5}{*}{Qwen2.5-7B}
    & GRPO-Full & 50.4 & 65.8 & 55.2 & 50.6 & 47.8 & 39.6 & 46.8 & 50.9 \\
  & & GRPO-Half & 49.8 & 63.2 & 57.4 & 42.4 & 42.6 & 28.2 & 29.6 & 44.7 \\
  & & SDGA-Anti & 22.6 & 37.8 & 20.4 & 21.8 & 26.8 & 5.8 & 21.4 & 22.4 \\
  & & SDGA-Auto & 58.2 & 73.6 & 71.6 & 56.2 & 55.8 & 45.4 & 47.4 & 58.3 \\
  & & \textbf{SDGA-Phase} & \textbf{60.8} & \textbf{76.4} & \textbf{73.6} & \textbf{59.2} & \textbf{56.4} & \textbf{51.6} & \textbf{57.8} & \textbf{62.3} \\
\midrule
\multirow{15}{*}{Search-R1}
  & \multirow{5}{*}{LLaMA-3.2-3B-Base}
    & GRPO-Full & 35.8 & 51.2 & 45.1 & 31.8 & 30.6 & 13.6 & 15.2 & 31.9 \\
  & & GRPO-Half & 33.1 & 47.4 & 41.2 & 27.8 & 27.6 & 10.1 & 12.7 & 28.6 \\
  & & SDGA-Anti & 18.7 & 22.8 & 16.9 & 14.8 & 15.7 & 9.2 & 8.3 & 15.2 \\
  & & \textbf{SDGA-Auto} & 42.9 & \textbf{59.2} & 53.8 & \textbf{38.8} & \textbf{38.0} & 20.8 & \textbf{23.4} & \textbf{39.6} \\
  & & SDGA-Phase & \textbf{43.2} & 58.7 & \textbf{54.1} & 38.3 & 37.6 & \textbf{21.1} & 22.9 & 39.4 \\
\cmidrule{2-11}
  & \multirow{5}{*}{Qwen2.5-3B}
    & GRPO-Full & 41.6 & \textbf{56.1} & 50.9 & 35.9 & 33.7 & 13.3 & 20.1 & 35.9 \\
  & & GRPO-Half & 36.7 & 52.1 & 45.3 & 31.5 & 32.5 & 13.4 & 20.9 & 33.2 \\
  & & SDGA-Anti & 22.9 & 27.3 & 20.7 & 17.9 & 18.3 & 11.1 & 10.1 & 18.3 \\
  & & SDGA-Auto & 42.9 & 52.8 & \textbf{55.9} & 34.8 & \textbf{37.3} & 17.1 & 21.4 & 37.5 \\
  & & \textbf{SDGA-Phase} & \textbf{44.8} & 52.1 & 55.4 & \textbf{41.2} & 37.1 & \textbf{23.4} & \textbf{32.3} & \textbf{40.9} \\
\cmidrule{2-11}
  & \multirow{5}{*}{Qwen2.5-7B}
    & GRPO-Full & 44.8 & 58.8 & 54.2 & 38.6 & 36.8 & 16.4 & 23.6 & 39.0 \\
  & & GRPO-Half & 42.6 & 55.4 & 56.2 & 31.8 & 32.4 & 10.2 & 12.4 & 34.4 \\
  & & SDGA-Anti & 20.8 & 34.6 & 19.2 & 19.4 & 24.2 & 8.4 & 16.8 & 20.5 \\
  & & SDGA-Auto & 48.2 & \textbf{59.2} & 60.8 & 40.2 & 41.6 & 20.8 & 24.6 & 42.2 \\
  & & \textbf{SDGA-Phase} & \textbf{51.4} & 58.6 & \textbf{62.4} & \textbf{47.2} & \textbf{43.2} & \textbf{28.6} & \textbf{38.4} & \textbf{47.1} \\
\bottomrule
\multicolumn{11}{l}{\small TQA = TriviaQA, 2Wiki = 2WikiMultiHopQA, BG = Bamboogle.}
\end{tabular}%
}
% \vspace{-7mm}
\end{table}

% \vspace{1mm}
\textbf{(1) CuSearch consistently improves performance across frameworks, model families, and model scales.}
As shown in Table~\ref{tab:main}, \textbf{SDGA-Phase} outperforms both \textbf{GRPO-Full} and \textbf{GRPO-Half} in nearly all cases across ZeroSearch and Search-R1.
On Qwen2.5-3B with ZeroSearch, SDGA-Phase improves average EM from 44.5 to 56.3 over GRPO-Full, a gain of \textbf{11.8} points.
Under the stricter update-budget-matched setting, SDGA-Phase also exceeds GRPO-Half by \textbf{17.0} points.
Similar gains are observed for Qwen2.5-7B, with improvements of \textbf{11.4} EM under ZeroSearch and \textbf{8.1} EM under Search-R1, and for LLaMA-3.2-3B-Base, confirming generalization across model families and scales.
While the largest gains are obtained under the ZeroSearch stochastic simulator, SDGA-Phase also yields consistent improvements under a real search API and chunk-based retrieval, as analyzed in Section~\ref{sec:condition} and Appendix~\ref{app:condition}.

% \vspace{1mm}
\textbf{(2) Explicit phase-adaptive curriculum outperforms implicit curriculum early in training.}
Both SDGA-Auto and SDGA-Phase constitute forms of curriculum learning: SDGA-Auto follows an implicit curriculum aligned with the model's training dynamics, as the greedy selection naturally concentrates on progressively deeper trajectories as the depth distribution shifts upward.
SDGA-Phase makes this curriculum explicit through a managed phase threshold.
Across settings, SDGA-Phase outperforms SDGA-Auto in nearly all cases, with the sole exception of LLaMA-3.2-3B-Base under Search-R1 where SDGA-Auto leads by 0.2 average EM.
The advantage of SDGA-Phase is most pronounced early in training: when deep-search buckets are sparsely populated, SDGA-Auto falls back to shallow buckets and its selection becomes indistinguishable from random, whereas SDGA-Phase targets the depth level the batch can reliably supply and advances only when deeper trajectories become sufficiently abundant.

\begin{figure}[t]
\centering
\includegraphics[width=\linewidth]{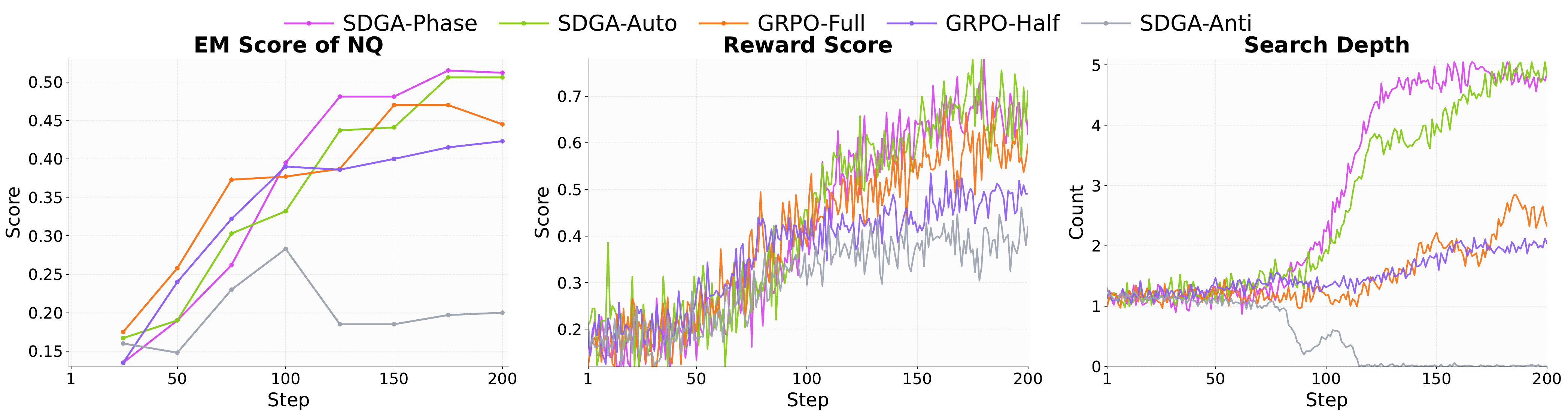}
\caption{Training dynamics on ZeroSearch with Qwen2.5-3B. Panels (a--c) show EM on NQ, mean reward, and average search depth per rollout. SDGA-Phase achieves the highest values on all three metrics by the end of training.}
\label{fig:training-dynamics}
% \vspace{-4mm}
\end{figure}

% \vspace{1mm}
\textbf{(3) The benefit depends on the retrieval environment.}
The gain of SDGA-Phase is larger on ZeroSearch than on Search-R1.
For Qwen2.5-3B, SDGA-Phase improves GRPO-Full by \textbf{11.8} EM on ZeroSearch and by \textbf{5.0} EM on Search-R1, raising the average from 35.9 to 40.9.
We analyze the contributions of search depth growth and retrieval stochasticity to this gap in Section~\ref{sec:condition} and Appendix~\ref{app:condition}.
Figure~\ref{fig:training-dynamics}(c) additionally shows that SDGA-Phase accelerates the growth of average search depth during training, indicating that the proposed allocation changes not only final task accuracy but also the model's retrieval behavior.

% \vspace{-4pt}
\subsection{Ablation Studies}

We conduct six ablation studies, with full results and tables in Appendix~\ref{app:ablation}.
\textbf{(1) Sampling signal and compute fairness.}
Random trajectory sampling, which shares the identical rollout generation and update cost as CuSearch, performs close to GRPO-Half but far below SDGA-Auto and SDGA-Phase.
This directly rules out a compute-based explanation: the gains of CuSearch stem from the depth-aware selection signal, not from training on fewer trajectories per step.
\textbf{(2) Robustness to $\Smax$.}
Both SDGA variants consistently outperform GRPO-Full across $\Smax \in \{3, 4, 5\}$, confirming robustness to the depth discretization granularity.
\textbf{(3) Direction of the signal.}
SDGA-Anti, which reverses the priority order, yields a drastic performance drop to 20.4 average EM compared to 52.9 for SDGA-Auto, confirming that the signal is directionally meaningful and that shallow-trajectory preference is actively harmful.
\textbf{(4) Transfer to a real search API.}
A policy trained with SDGA-Phase under ZeroSearch retains a \textbf{5.1} EM advantage over GRPO-Full when evaluated with a real SERPER API, confirming that learned retrieval behaviors transfer beyond the simulator.
\textbf{(5) Reward-based trajectory selection.}
Top-K Reward achieves 41.8 average EM, close to Random (41.9) and \textbf{14.5} points below SDGA-Phase. Notably, this score also falls below GRPO-Full, which achieves 44.5 EM. This suggests that concentrating updates on the highest-reward trajectories can hurt training by introducing selection bias, confirming that reward magnitude alone does not provide an effective selection signal and that search depth captures a structurally distinct and more informative criterion.
\textbf{(6) Prompt-level adaptive baselines.}
SDGA-Phase outperforms DAPO by \textbf{5.9} EM and AR3PO by \textbf{6.7} EM under matched compute, indicating that rollout-level structural selection is both distinct from and more effective than prompt-level filtering for agentic RAG.

% \vspace{-4pt}
\subsection{In-Depth Analysis}
\label{sec:condition}

% \vspace{2mm}
\textbf{(C1) Search depth grows during training.}
Under both ZeroSearch and Search-R1, answer-correctness reward alone drives a monotonic increase in per-trajectory search counts, providing a progressively larger supply of deep-search trajectories. The growth is more pronounced under ZeroSearch, partially explaining the larger gains in that setting.

% \vspace{2mm}
\textbf{(C2) Retrieval-environment stochasticity amplifies the gain.}
In more stochastic environments, diverse search decisions produce varied evidence, making each additional search step more informative. In deterministic environments, repeated queries return overlapping results, reducing the marginal value of deeper trajectories. ZeroSearch uses an LLM-simulated retriever that produces varied results across queries. Search-R1 uses a fixed offline corpus where repeated or similar queries tend to return overlapping documents, reducing the marginal benefit of additional search steps.
Table~\ref{tab:condition} shows gains of \textbf{11.8} EM under ZeroSearch, \textbf{5.1} EM with a real search API, and \textbf{5.0} EM under Search-R1, consistent with C1 as the primary driver and C2 as a secondary amplifier in high-stochasticity settings. A detailed analysis is provided in Appendix~\ref{app:condition}.

Taken together, these results confirm that search depth is a reliable and annotation-free signal for selecting informative trajectories. Allocating the update budget toward deeper-search rollouts yields consistent improvements over both uniform sampling and prompt-level adaptive methods across diverse training conditions, providing direct evidence for the central claim of this work.
\section{Conclusion}

Per-trajectory search depth is a reliable, annotation-free proxy for retrieval supervision density in RLVR-based agentic RAG training. Search counts increase monotonically over training, and the within-batch depth distribution shifts toward higher values, deepening the informativeness gap that uniform rollout usage fails to exploit.
To address this, we propose \textbf{CuSearch}, built on the \textbf{SDGA} operator. \textbf{SDGA-Auto} always targets the deepest available trajectories in the current batch, yielding an implicit training-aligned curriculum as the depth distribution shifts upward. \textbf{SDGA-Phase} explicitly advances the curriculum threshold as deeper trajectories become sufficiently abundant.
Consistent gains across ZeroSearch and Search-R1 confirm that depth-aware rollout allocation improves retrieval sub-policy learning across model families and retrieval frameworks. Gain magnitude is governed primarily by search depth growth during training and amplified by retrieval-environment stochasticity.

\paragraph{Limitations.}
The method yields smaller gains when search depth growth is limited or when the retrieval environment has low stochasticity.
It uses search depth as a single scalar feature and does not condition on reward quality, though within-group advantage normalization mitigates this by suppressing the gradient contribution of low-reward trajectories even when they are selected.
Combining search depth with reward-derived signals and extending CuSearch to other agentic behaviors such as tool use and code execution are natural directions for future work.

% ── References ───────────────────────────────────────────────────────────────
\bibliography{colm2026_conference}
\bibliographystyle{colm2026_conference}

\clearpage
\appendix
\section{Training Details}
\label{app:training-details}

\subsection{Prompt Template}
\label{app:prompt-template}

We follow the prompt template used in Search-R1 and ZeroSearch~\citep{jin2025searchr1,shen2025zerosearch}. 
The template used during training and inference is shown below:

\begin{figure}[h]
\centering
\setlength{\fboxsep}{0pt}
\fbox{
\parbox{0.97\linewidth}{
\vspace{0.6em}
\noindent
Answer the given question. You must conduct reasoning inside {\color{blue}\texttt{<think>}} and {\color{blue}\texttt{</think>}} first every time you get new information. After reasoning, if you find you lack some knowledge, you can call a search engine by {\color{cyan}\texttt{<search>}} query {\color{cyan}\texttt{</search>}}, and it will return the top searched results between {\color{brown}\texttt{<information>}} and {\color{brown}\texttt{</information>}}. You can search as many times as you want. If you find no further external knowledge needed, you can directly provide the answer inside {\color{red}\texttt{<answer>}} and {\color{red}\texttt{</answer>}} without detailed illustrations. For example, {\color{red}\texttt{<answer>}} Beijing {\color{red}\texttt{</answer>}}. Question:
\vspace{0.6em}
}
}
\caption{Training template. The question is appended to the end during training and inference.}
\label{fig:prompt-template}
\end{figure}

\subsection{Reward Design}
\label{app:reward-design}

We follow the reward design used in Search-R1 and adopt a rule-based reward that combines answer quality with format compliance~\citep{jin2025searchr1}. 
Following ZeroSearch~\citep{shen2025zerosearch}, we use the F1 score rather than exact-match accuracy as the answer-quality signal. 
Specifically, the reward for a trajectory $\tau$ is defined as
\begin{equation}
  r(\tau) = \mathrm{F1}(\tau)\,(1-\lambda_f) + \lambda_f\,\rfmt(\tau),
  \label{eq:reward}
\end{equation}
where $\mathrm{F1}(\tau) \in [0,1]$ denotes the F1 score between the extracted final answer and the ground-truth answer, and $\rfmt(\tau) \in \{0,1\}$ is a binary indicator of format validity. 

A trajectory is considered format-valid if it follows the prescribed agent format, with reasoning enclosed by \texttt{<think>...</think>}, search queries enclosed by \texttt{<search>...</search>}, and the final answer enclosed by \texttt{<answer>...</answer>}. 
This design encourages the model not only to produce high-quality answers but also to adhere to the interaction protocol required for search-engine invocation.

Following the empirical findings of Search-R1, we do not use an additional intermediate retrieval reward. 
Instead, we use the answer-quality reward together with the format-compliance signal throughout all experiments.

\begin{table}[h]
\centering
\caption{Main training hyperparameters used in our experiments. Unless otherwise specified, all settings follow the default configuration of ZeroSearch.}
\label{tab:hyperparameters}
\begin{tabular}{lc}
\toprule
Hyperparameter & Value \\
\midrule
Number of prompts per step $N$ & 64 \\
Rollouts per prompt $G$ & 3 \\
Selected trajectories per update $K$ & 96 \\
Mini-batch size $M$ & 32 \\
Maximum search count $\Smax$ & 5 \\
Learning rate & $1 \times 10^{-6}$ \\
KL penalty weight $\beta$ & 0.001 \\
Format reward weight $\lambda_f$ & 0.2 \\
\bottomrule
\end{tabular}
\end{table}

\subsection{Training Hyperparameters}
\label{app:hyperparameters}

Unless otherwise specified, all training hyperparameters follow the default configuration of ZeroSearch and are kept fixed across all methods.
CuSearch does not introduce additional trainable components or modify the underlying policy optimization procedure.
The complete hyperparameter values are listed in Table~\ref{tab:hyperparameters}.

% ─────────────────────────────────────────────────────────────────────────────
\section{CuSearch Training Loop}
\label{app:cusearch-loop}
% ─────────────────────────────────────────────────────────────────────────────

Algorithm~\ref{alg:cusearch} summarizes CuSearch within a standard VERL/GRPO training pipeline.
At each step, the policy generates $G$ rollouts per prompt. CuSearch parses search counts, applies SDGA, and retains only $K$ trajectories for reward computation and gradient updates.

\begin{algorithm}[H]
\SetAlgoLined
\caption{CuSearch Training Loop}
\label{alg:cusearch}
\KwIn{Prompt pool $P$; policies $\pi_\theta$, $\pi_{\mathrm{ref}}$; hyperparameters $N,G,K,M,\beta,\eta$}
\KwOut{Trained policy $\pi_\theta$}

\For{training step $t = 1,2,\ldots$}{
  \tcp{Phase 1: Rollout generation}
  $\{x_i\}_{i=1}^{N} \leftarrow \mathrm{sample}(P,N)$\;
  \For{$i=1,\ldots,N$}{
    \For{$g=1,\ldots,G$}{
      $\tau_{i,g} \leftarrow \mathrm{agent\_loop}(\pi_\theta, x_i)$\;
    }
  }
  $\mathcal{T} \leftarrow \{\tau_{i,g}\}_{i=1,g=1}^{N,G}$\;

  \tcp{Phase 2: Search-count parsing and trajectory selection}
  compute $\mathrm{cnt}(\tau)$ for each $\tau \in \mathcal{T}$\;
  construct buckets $\{\mathcal{B}_s\}_{s=0}^{S_{\max}}$ and capacities $\{c_s\}_{s=0}^{S_{\max}}$\;
  compute target allocations and priorities according to the chosen SDGA variant\;
  obtain feasible allocations $\{f_s\}_{s=0}^{S_{\max}} \leftarrow \mathrm{SDGA}(\{c_s\}, \{t_s\}, \{p_s\})$\;
  sample $f_s$ trajectories from each bucket $\mathcal{B}_s$ and form $T$\;

  \tcp{Phase 3: Reward and advantage computation}
  \ForEach{$\tau \in T$}{
    $r(\tau) \leftarrow \text{Eq.~\eqref{eq:reward}}$\;
  }
  compute advantages on the selected set $T$\;

  \tcp{Phase 4: Policy update}
  \ForEach{mini-batch $B \subseteq T$ of size $M$}{
    $\theta \leftarrow \theta - \eta \nabla_\theta \mathcal{L}_{\mathrm{CuSearch}}(\theta; B, \pi_{\mathrm{ref}}, \beta)$\;
  }
}
\Return $\pi_\theta$\;
\end{algorithm}

% ─────────────────────────────────────────────────────────────────────────────
\section{Detailed Instantiations of SDGA}
\label{app:sdga}

This appendix provides the detailed definitions of the two instantiations of Search-Depth Greedy Allocation (SDGA): \textbf{SDGA-Auto} and \textbf{SDGA-Phase}.

Given the rollout set $\mathcal{T}$, we partition trajectories into $S_{\max}+1$ buckets according to search count:
\begin{equation}
  \mathcal{B}_s = \{\tau \in \mathcal{T} : \mathrm{cnt}(\tau) = s\},
  \quad s \in \{0,1,\ldots,S_{\max}\},
\end{equation}
where $c_s = |\mathcal{B}_s|$ and $\sum_s c_s = |\mathcal{T}|$. Let $K = |T|$ denote the total selection budget. Each SDGA variant is specified by a target allocation $\{t_s\}$ and a priority assignment $\{p_s\}$, which are converted into a feasible allocation under the bucket-capacity constraints.

\subsection{Additional Details of SDGA}
\label{app:sdga-details}

\begin{algorithm}[H]
\SetAlgoLined
\caption{Search-Depth Greedy Allocation (SDGA): Detailed Version}
\label{alg:sdga_appendix}
\KwIn{
Capacities $\{c_s\}_{s=0}^{S_{\max}}$;
target allocations $\{t_s\}_{s=0}^{S_{\max}}$ with $\sum_{s=0}^{S_{\max}} t_s = K$;
priorities $\{p_s\}_{s=0}^{S_{\max}}$, where smaller values indicate higher priority
}
\KwOut{Feasible allocations $\{f_s\}_{s=0}^{S_{\max}}$ satisfying $0 \le f_s \le c_s$}

$f_s \leftarrow t_s$ for all $s$\;
$\sigma \leftarrow \mathrm{argsort}(\{p_s\})$ \tcp*{priority order: high to low}

\ForEach{$s$ in order of $\sigma$}{
    \If{$f_s \le c_s$}{
        \textbf{continue}\;
    }

    $d \leftarrow f_s - c_s$ \tcp*{overflow of bucket $s$}
    $f_s \leftarrow c_s$\;

    \tcp{Step 1: redistribute overflow to lower-priority buckets}
    \ForEach{$r$ after $s$ in $\sigma$}{
        \If{$d = 0$}{
            \textbf{break}\;
        }

        $\mathrm{spare} \leftarrow c_r - f_r$\;
        \If{$\mathrm{spare} > 0$}{
            $\delta \leftarrow \min(\mathrm{spare}, d)$\;
            $f_r \leftarrow f_r + \delta$\;
            $d \leftarrow d - \delta$\;
        }
    }

    \tcp{Step 2: if needed, redistribute remaining overflow to higher-priority buckets}
    \If{$d > 0$}{
        \ForEach{$r$ before $s$ in $\sigma$, in reverse order}{
            \If{$d = 0$}{
                \textbf{break}\;
            }

            $\mathrm{spare} \leftarrow c_r - f_r$\;
            \If{$\mathrm{spare} > 0$}{
                $\delta \leftarrow \min(\mathrm{spare}, d)$\;
                $f_r \leftarrow f_r + \delta$\;
                $d \leftarrow d - \delta$\;
            }
        }
    }
}
\Return $\{f_s\}_{s=0}^{S_{\max}}$\;
\end{algorithm}

Algorithm~\ref{alg:sdga_appendix} gives the detailed procedure of SDGA. 
Given bucket capacities $\{c_s\}_{s=0}^{S_{\max}}$, target allocations $\{t_s\}_{s=0}^{S_{\max}}$, and priorities $\{p_s\}_{s=0}^{S_{\max}}$, SDGA transforms the target allocation into a feasible allocation that satisfies the capacity constraint of every bucket. 
The algorithm is initialized with the target allocation and processes buckets in the order induced by the priority values, where smaller values indicate higher priority.

When the allocation of a bucket exceeds its capacity, the excess is treated as overflow and redistributed greedily. 
For an over-allocated bucket, SDGA first transfers the overflow to the lower-priority buckets that follow later in the priority order.
If overflow remains, the algorithm then redistributes it to the higher-priority buckets that precede it in the priority order, traversed in reverse from lowest to highest rank. 
This repair procedure preserves the priority structure as much as possible while enforcing feasibility.

The resulting allocation $\{f_s\}_{s=0}^{S_{\max}}$ satisfies $0 \le f_s \le c_s$ for all $s$. 
Moreover, when the total target allocation does not exceed the total available capacity, that is $\sum_s t_s \le \sum_s c_s$, the total allocation is preserved, so that $\sum_s f_s = \sum_s t_s$. 
In our method, SDGA serves as a generic repair operator after target bucket counts are specified by a search-aware allocation policy.

\subsection{SDGA-Auto}
\label{app:sdga-auto}

SDGA-Auto is a stateless instantiation of SDGA that always prioritizes trajectories with larger search counts. 
Its target allocation places the entire selection budget on the highest-search bucket:
\begin{equation}
  t_s =
  \begin{cases}
    K & \text{if } s = S_{\max},\\
    0 & \text{otherwise},
  \end{cases}
\end{equation}
where $K$ denotes the number of selected trajectories per update. 
The priority assignment is defined in descending order of search count:
\begin{equation}
  p_s = S_{\max} - s + 1,
\end{equation}
so that larger search counts correspond to higher priority. 
Equivalently, the induced priority order is
\begin{equation}
  S_{\max} \to S_{\max}-1 \to \cdots \to 0.
\end{equation}

Under this setting, SDGA first attempts to allocate the full budget to the bucket with the largest search count. 
If that bucket does not contain enough trajectories, the remaining budget is redistributed greedily to buckets with progressively smaller search counts according to the SDGA repair procedure.

For example, when $S_{\max}=5$, the corresponding target allocation and priorities can be written as
\begin{equation}
  \texttt{target\_lens} = [0,0,0,0,0,K],
  \qquad
  \texttt{priorities} = [6,5,4,3,2,1].
\end{equation}

\subsection{SDGA-Phase}
\label{app:sdga-phase}

SDGA-Phase is a phase-adaptive instantiation of SDGA. 
It maintains a global phase variable $k \in \{0,1,\ldots,S_{\max}-1\}$, initialized to $0$ and constrained to be monotonically non-decreasing. 
At phase $k$, the target allocation places the full selection budget on bucket $k+1$:
\begin{equation}
  t_s =
  \begin{cases}
    K & \text{if } s = k+1,\\
    0 & \text{otherwise},
  \end{cases}
\end{equation}
where $K$ denotes the number of selected trajectories per update. 
The corresponding priority assignment gives the highest priority to the current target bucket, then to buckets with larger search counts, and finally to buckets with smaller search counts. 
Formally,
\begin{equation}
  p(s,k) =
  \begin{cases}
    s-k & \text{if } s \ge k+1,\\[2pt]
    S_{\max}+1-s & \text{if } s < k+1.
  \end{cases}
  \label{eq:appendix-phase-priority}
\end{equation}
This induces the priority order
\begin{equation}
  k+1 \to k+2 \to \cdots \to S_{\max} \to k \to k-1 \to \cdots \to 0.
\end{equation}

Under this ordering, SDGA-Phase first attempts to allocate trajectories from the current target bucket. 
If that bucket does not contain enough trajectories, the remaining budget is first redistributed to buckets with larger search counts and only then to buckets with smaller search counts. 
This mechanism induces a curriculum over search depth, allowing trajectory selection to shift progressively toward deeper retrieval behavior as training advances.

\subsubsection{Phase Transition}
The phase is advanced from $k$ to $k+1$ when
\begin{equation}
  \sum_{s>k+1} c_s \ge K,
\end{equation}
that is, when buckets strictly above the current target bucket can already satisfy the full selection budget.
Once advanced, the phase never decreases, which prevents oscillation and enforces a monotonic progression of the search-depth curriculum.

\subsubsection{Concrete Examples}
When $S_{\max}=5$, the target allocations and priorities for each phase are as follows.
At phase $0$,
\begin{equation}
  \texttt{target\_lens} = [0,K,0,0,0,0],
  \qquad
  \texttt{priorities} = [6,1,2,3,4,5].
\end{equation}
At phase $1$,
\begin{equation}
  \texttt{target\_lens} = [0,0,K,0,0,0],
  \qquad
  \texttt{priorities} = [6,5,1,2,3,4].
\end{equation}
At phase $2$,
\begin{equation}
  \texttt{target\_lens} = [0,0,0,K,0,0],
  \qquad
  \texttt{priorities} = [6,5,4,1,2,3].
\end{equation}
At phase $3$,
\begin{equation}
  \texttt{target\_lens} = [0,0,0,0,K,0],
  \qquad
  \texttt{priorities} = [6,5,4,3,1,2].
\end{equation}
At phase $4$,
\begin{equation}
  \texttt{target\_lens} = [0,0,0,0,0,K],
  \qquad
  \texttt{priorities} = [6,5,4,3,2,1].
\end{equation}
Note that at phase $4$, the target allocation and priority ordering coincide with SDGA-Auto, as the curriculum has reached the deepest available search depth.
For instance, when $k=1$, the induced priority order is
\begin{equation}
  2 \to 3 \to 4 \to 5 \to 1 \to 0.
\end{equation}
In this case, the allocator first targets search count $2$, then progressively shifts toward larger search counts, and falls back to smaller ones only when necessary.

\subsection{SDGA-Anti}
\label{app:sdga-anti}

SDGA-Anti is an anti-curriculum instantiation of SDGA that reverses the priority order of SDGA-Auto, deliberately selecting the shallowest-search trajectories.
It places the entire selection budget on the lowest-search bucket:
\begin{equation}
  t_s =
  \begin{cases}
    K & \text{if } s = 0,\\
    0 & \text{otherwise},
  \end{cases}
\end{equation}
where $K$ denotes the number of selected trajectories per update. 
The priority assignment is defined in ascending order of search count:
\begin{equation}
  p_s = s + 1,
\end{equation}
so that smaller search counts correspond to higher priority. 
Equivalently, the induced priority order is
\begin{equation}
  0 \to 1 \to \cdots \to S_{\max}.
\end{equation}

Under this setting, SDGA first attempts to allocate the full budget to the bucket with the smallest search count. 
If that bucket does not contain enough trajectories, the remaining budget is redistributed greedily to buckets with progressively larger search counts according to the SDGA repair procedure.

For example, when $S_{\max}=5$, the corresponding target allocation and priorities can be written as
\begin{equation}
  \texttt{target\_lens} = [K,0,0,0,0,0],
  \qquad
  \texttt{priorities} = [1,2,3,4,5,6].
\end{equation}

% ─────────────────────────────────────────────────────────────────────────────
\section{Ablation Studies}
\label{app:ablation}
% ─────────────────────────────────────────────────────────────────────────────

\subsection{Ablation 1: Is Search Depth Itself a Useful Sampling Signal?}
\label{app:ablation-signal}

\begin{table}[h]
\caption{Ablation on trajectory selection signals under the ZeroSearch framework (simulator evaluation, Qwen2.5-3B, EM, \%). Random and Top-K Reward share the same rollout generation and update cost as CuSearch.}
\label{tab:random}
\centering
\small
\begin{tabular}{lcccccccc}
\toprule
Method & NQ & TQA & PopQA & HotpotQA & 2Wiki & MuSiQue & BG & Avg. \\
\midrule
GRPO-Full      & 44.3 & 60.7 & 48.1 & 44.5 & 41.2 & 32.7 & 39.8 & 44.5 \\
GRPO-Half      & 44.1 & 58.5 & 52.3 & 36.9 & 36.6 & 22.3 & 24.1 & 39.3 \\
Random         & 42.6 & 58.8 & 56.0 & 35.8 & 37.8 & 22.8 & 39.2 & 41.9 \\
Top-K Reward   & 43.8 & 60.2 & 49.8 & 36.9 & 38.6 & 24.8 & 38.4 & 41.8 \\
\midrule
SDGA-Auto        & 53.4 & 67.3 & \textbf{68.0} & 51.5 & \textbf{51.8} & 39.4 & 39.1 & 52.9 \\
\textbf{SDGA-Phase}  & \textbf{55.1} & \textbf{72.4} & 67.9 & \textbf{52.8} & 49.5 & \textbf{44.9} & \textbf{51.4} & \textbf{56.3} \\
\bottomrule
\end{tabular}
\end{table}

Random achieves 41.9 average EM, substantially below SDGA-Auto at 52.9 and SDGA-Phase at 56.3, while remaining close to GRPO-Half at 39.3.
Because Random uniformly samples $K$ trajectories from the full rollout pool, it preserves the original within-batch depth distribution and does not preferentially allocate updates to deeper-search trajectories.
Its near-parity with GRPO-Half confirms that simply reducing the number of trajectory updates is not sufficient.
The gap between Random and SDGA-Auto therefore isolates the value of the sampling signal itself: both use the same update budget $K$, but only SDGA-Auto concentrates that budget on trajectories with more retrieval decision points.
This result directly supports our claim that search depth is a useful structural proxy for trajectory informativeness.

\subsection{Ablation 2: Robustness to $\Smax$}
\label{app:ablation-smax}

\begin{table}[h]
\caption{Performance under different $\Smax$ settings (Qwen2.5-3B, ZeroSearch simulator evaluation, EM, \%).}
\label{tab:smax}
\centering
\small
\begin{tabular}{clcccccccc}
\toprule
$\Smax$ & Method & NQ & TQA & PopQA & HotpotQA & 2Wiki & MuSiQue & BG & Avg. \\
\midrule
\multirow{4}{*}{5}
  & GRPO-Full  & 44.3 & 60.7 & 48.1 & 44.5 & 41.2 & 32.7 & 39.8 & 44.5 \\
  & GRPO-Half  & 44.1 & 58.5 & 52.3 & 36.9 & 36.6 & 22.3 & 24.1 & 39.3 \\
  & SDGA-Auto        & 53.4 & 67.3 & \textbf{68.0} & 51.5 & \textbf{51.8} & 39.4 & 39.1 & 52.9 \\
  & \textbf{SDGA-Phase}  & \textbf{55.1} & \textbf{72.4} & 67.9 & \textbf{52.8} & 49.5 & \textbf{44.9} & \textbf{51.4} & \textbf{56.3} \\
\midrule
\multirow{4}{*}{4}
  & GRPO-Full  & 44.0 & 59.6 & 52.8 & 39.8 & 37.6 & 26.2 & 30.4 & 41.5 \\
  & GRPO-Half  & 42.4 & 57.2 & 48.6 & 38.6 & 37.2 & 25.2 & 28.8 & 39.7 \\
  & SDGA-Auto        & \textbf{51.8} & \textbf{66.0} & 55.8 & 42.2 & 40.2 & 32.2 & 34.4 & 46.1 \\
  & \textbf{SDGA-Phase}  & 46.2 & 64.0 & \textbf{57.0} & \textbf{43.4} & \textbf{43.6} & \textbf{33.6} & \textbf{43.2} & \textbf{47.3} \\
\midrule
\multirow{4}{*}{3}
  & GRPO-Full  & 43.2 & 62.0 & 54.0 & 40.0 & 41.8 & 24.6 & \textbf{40.0} & 43.7 \\
  & GRPO-Half  & 42.8 & 62.6 & 54.6 & 36.6 & 40.4 & 26.8 & 24.0 & 41.1 \\
  & SDGA-Auto        & \textbf{48.8} & \textbf{67.2} & \textbf{60.8} & 41.0 & \textbf{45.8} & 31.6 & 29.6 & \textbf{46.4} \\
  & \textbf{SDGA-Phase}  & 41.8 & 60.0 & 57.6 & \textbf{47.0} & 43.4 & \textbf{34.4} & 32.0 & 45.2 \\
\bottomrule
\end{tabular}
\end{table}

As shown in Table~\ref{tab:smax}, both SDGA-Auto and SDGA-Phase consistently outperform GRPO-Full across $\Smax \in \{3, 4, 5\}$.
This indicates that the benefit of CuSearch is robust to the discretization granularity used for search depth.
Increasing $\Smax$ yields finer priority stratification and generally better performance, suggesting that preserving more depth information helps distinguish informative trajectories more precisely.

\subsection{Ablation 3: Does the Direction of the Signal Matter?}
\label{app:ablation-direction}

\begin{table}[t]
\caption{Ablation on sampling priority direction under ZeroSearch (simulator evaluation, Qwen2.5-3B, EM, \%).}
\label{tab:direction}
\centering
\small
\resizebox{\linewidth}{!}{%
\begin{tabular}{llcccccccc}
\toprule
Method & Priority Direction & NQ & TQA & PopQA & HotpotQA & 2Wiki & MuSiQue & BG & Avg. \\
\midrule
SDGA-Anti  & Low-search first $\downarrow$ & 20.2 & 35.3 & 17.7 & 19.9 & 24.7 & 4.3  & 20.9 & 20.4 \\
Random   & Uniform                       & 42.6 & 58.8 & 56.0 & 35.8 & 37.8 & 22.8 & \textbf{39.2} & 41.9 \\
\textbf{SDGA-Auto} & High-search first $\uparrow$ & \textbf{53.4} & \textbf{67.3} & \textbf{68.0} & \textbf{51.5} & \textbf{51.8} & \textbf{39.4} & 39.1 & \textbf{52.9} \\
\bottomrule
\end{tabular}
}
\end{table}

Table~\ref{tab:direction} shows results on ZeroSearch with Qwen2.5-3B, where the methods follow a strict ordering:
SDGA-Anti (20.4) $<$ Random (41.9) $<$ SDGA-Auto (52.9).
The total spread is 32.5 EM points, and SDGA-Anti falls to 4.3 on MuSiQue.
This result shows that the signal is not merely correlated with performance but directionally meaningful:
prioritizing shallower trajectories is not just less effective than prioritizing deeper ones, but can be actively harmful.
This strongly supports our interpretation of search depth as an informativeness-oriented supervision signal.

\subsection{Ablation 4: Generalization to a Real Search Engine (SERPER)}
\label{app:ablation-serper}

\begin{table}[h]
\caption{Transfer performance on SERPER real-search evaluation using ZeroSearch-trained Qwen2.5-3B models (EM, \%).}
\label{tab:serper}
\centering
\small
\begin{tabular}{lcccccccc}
\toprule
Method & NQ & TQA & PopQA & HotpotQA & 2Wiki & MuSiQue & BG & Avg. \\
% \midrule
% \multicolumn{9}{l}{\textit{LLaMA-3.2-3B-Base}} \\
% \midrule
% GRPO-Full  & --- & --- & --- & --- & --- & --- & --- & --- \\
% GRPO-Half  & --- & --- & --- & --- & --- & --- & --- & --- \\
% \textbf{SDGA-Auto}        & --- & --- & --- & --- & --- & --- & --- & --- \\
% \textbf{SDGA-Phase}  & --- & --- & --- & --- & --- & --- & --- & --- \\
% \midrule
\multicolumn{9}{l}{\textit{Qwen2.5-3B}} \\
\midrule
GRPO-Full  & 43.6 & 62.2 & 55.2 & 36.2 & 38.6 & 20.6 & 24.0 & 40.1 \\
GRPO-Half  & 43.2 & 64.2 & 57.8 & 35.4 & 34.6 & 17.4 & 23.2 & 39.4 \\
\textbf{SDGA-Auto}        & 46.6 & \textbf{67.2} & 62.2 & 38.0 & 41.0 & 20.4 & 24.8 & 42.9 \\
\textbf{SDGA-Phase}  & \textbf{48.0} & 66.2 & \textbf{64.6} & \textbf{40.4} & \textbf{43.0} & \textbf{22.2} & \textbf{32.0} & \textbf{45.2} \\
\bottomrule
\end{tabular}
\end{table}

A policy trained with SDGA-Phase under ZeroSearch still outperforms GRPO-Full by \textbf{5.1} EM points, raising the average from 40.1 to 45.2, when evaluated with a real search API (SERPER).
The gain is smaller than the \textbf{11.8}-point improvement observed under simulator evaluation, indicating that some of the benefit is tied to the simulator's retrieval distribution.
Nevertheless, the remaining margin is still substantial, showing that the search behaviors encouraged by CuSearch transfer beyond the training environment.

\subsection{Ablation 5: Reward-Based Trajectory Selection}
\label{app:ablation-reward}

To test whether search depth provides a selection signal beyond reward magnitude, we compare against \textbf{Top-K Reward}, which selects the $K$ trajectories with the highest individual rewards from the rollout pool $\mathcal{T}$.
Results are included in Table~\ref{tab:random} alongside the other selection-signal baselines.

Top-K Reward achieves 41.8 average EM, close to Random (41.9) and substantially below SDGA-Auto (52.9) and SDGA-Phase (56.3).
This result is consistent with a structural observation: high-reward trajectories are not necessarily deep-search trajectories.
A model that answers correctly without issuing any retrieval calls receives a high reward but contributes zero retrieval gradient coverage.
Preferring such trajectories concentrates the update budget on answer-generation behavior while depriving the retrieval sub-policy of direct supervision.
In contrast, search-depth-based selection explicitly targets trajectories that provide the most retrieval decision points per update, regardless of whether those trajectories receive high or low rewards.
Advantage normalization within GRPO then handles the reward signal: low-reward deep-search trajectories receive suppressed gradient weights, while high-reward deep-search trajectories receive amplified weights.
Search depth and reward therefore operate at complementary levels, and their combination within SDGA produces a more effective training signal than either alone.

\subsection{Ablation 6: Comparison with Prompt-Level Adaptive Methods}
\label{app:ablation-prompt}

\begin{table}[h]
\caption{Comparison with prompt-level adaptive methods under ZeroSearch (simulator evaluation, Qwen2.5-3B, EM, \%). All methods use the same total rollout budget ($N \cdot G = 192$).}
\label{tab:prompt-level}
\centering
\small
\resizebox{\linewidth}{!}{
\begin{tabular}{lcccccccc}
\toprule
Method & NQ & TQA & PopQA & HotpotQA & 2Wiki & MuSiQue & BG & Avg. \\
\midrule
GRPO-Full & 44.3 & 60.7 & 48.1 & 44.5 & 41.2 & 32.7 & 39.8 & 44.5 \\
DAPO~\citep{yu2025dapo} & 49.1 & 66.5 & 55.3 & 50.2 & 47.1 & 38.7 & 45.9 & 50.4 \\
AR3PO~\citep{zhang2025ar3po} & 48.6 & 65.2 & 54.7 & 49.5 & 46.4 & 37.9 & 45.1 & 49.6 \\
\textbf{SDGA-Phase (Ours)} & \textbf{55.1} & \textbf{72.4} & \textbf{67.9} & \textbf{52.8} & \textbf{49.5} & \textbf{44.9} & \textbf{51.4} & \textbf{56.3} \\
\bottomrule
\end{tabular}
}
\end{table}

\textbf{Implementation.}
For DAPO, we implement only the dynamic sampling component: at each training step, prompt groups in which all $G$ rollouts receive the same reward are removed before the policy update, as these groups produce zero group-normalized advantage and contribute no learning signal.
All other hyperparameters, including the loss formulation, clipping coefficient $\varepsilon$, and KL penalty weight $\beta$, are kept identical to the GRPO baseline.
For AR3PO, we implement the adaptive rollout allocation: we track the per-prompt pass rate (fraction of $G$ rollouts receiving a positive reward) over a sliding window, and reallocate the rollout budget to favor prompts with intermediate pass rates.
Prompts whose recent pass rate falls near $0$ or $1$ receive fewer rollouts, while prompts with pass rates in the intermediate range receive more.
The total rollout budget per step is held fixed at $N \cdot G = 192$ across all prompt-level baselines and CuSearch to ensure a fair comparison.

As shown in Table~\ref{tab:prompt-level}, DAPO and AR3PO improve average EM over GRPO-Full by 5.9 and 5.1 points respectively, whereas SDGA-Phase achieves an 11.8-point gain under the same rollout budget.
This comparison highlights the difference between the two levels of adaptation.
Prompt-level methods decide \emph{which prompts} deserve more training signal, while our approach decides \emph{which trajectories within a batch} carry more informative retrieval supervision.
Our results suggest that, for agentic RAG, exploiting rollout-level structural heterogeneity yields substantially larger gains than prompt-level filtering alone.
Because the two operate at different levels, they are potentially complementary.

% ─────────────────────────────────────────────────────────────────────────────
\section{In-Depth Analysis: Effectiveness Conditions}
\label{app:condition}
% ─────────────────────────────────────────────────────────────────────────────

\begin{table}[h]
\caption{SDGA-Phase gain over GRPO-Full ($\Delta$EM, \%) across retrieval frameworks and corresponding retrieval environment stochasticity (condition C2). Results on Qwen2.5-3B.}
\label{tab:condition}
\centering
\small
\begin{tabular}{lcccc}
\toprule
Retrieval Framework & C2 (Env.\ Stochasticity) & GRPO-Full Avg. & SDGA-Phase Avg. & $\Delta$EM \\
\midrule
ZeroSearch (Simulator) & High (LLM simulator)   & 44.5 & 56.3 & $+11.8$ \\
ZeroSearch (Real API)  & Medium (web search)    & 40.1 & 45.2 & $+5.1$  \\
Search-R1              & Low (chunk retrieval)  & 35.9 & 40.9 & $+5.0$  \\
\bottomrule
\end{tabular}
\end{table}

To better understand when CuSearch is effective, we analyze the method through the lens of two conditions.

\textbf{Condition C1: Search depth grows during training.}
CuSearch reallocates the within-batch update budget toward deeper-search trajectories.
This mechanism is only meaningful if deeper-search trajectories emerge with non-trivial frequency.
Figure~\ref{fig:search-dist}(a) shows that under ZeroSearch, answer-correctness reward alone drives a monotonic increase in average search count throughout training.
The same trend holds under Search-R1, where average valid searches per rollout rise from approximately 1.0 to 1.3 and stabilize around 1.2.
This confirms that C1 is a general property of RLVR-based agentic RAG training, not specific to a particular retrieval framework.
However, the magnitude of growth differs: ZeroSearch exhibits a steeper and more sustained increase, while Search-R1 reaches a lower plateau.
This difference in growth rate partially explains why CuSearch yields larger gains under ZeroSearch.
When search depth growth is absent, the within-batch depth distribution remains concentrated at low counts, and search-depth-based allocation collapses toward random selection.

\textbf{Condition C2: Retrieval-environment stochasticity amplifies the gain.}
Beyond the supply of deep-search trajectories, the value of each additional search step depends on how much new evidence it contributes.
In a more stochastic retrieval environment, different search decisions are more likely to produce diverse observations, so deeper trajectories carry more unique supervision signal per step.
In a more deterministic environment, repeated or similar queries often return overlapping evidence, reducing the marginal utility of additional search steps.
Table~\ref{tab:condition} is consistent with this account: the gain of SDGA-Phase over GRPO-Full is \textbf{11.8} EM in the stochastic ZeroSearch simulator, \textbf{5.1} EM with a real search API, and \textbf{5.0} EM in chunk-based Search-R1.
The simulator yields the largest gain, while the two lower-stochasticity settings yield smaller and comparable improvements.
Importantly, the near-identical gains under the real API and Search-R1 suggest that C1 is the primary driver of the large ZeroSearch advantage: ZeroSearch exhibits substantially stronger search depth growth, which provides a larger selectable budget of deep trajectories at each training step.
C2 acts as a secondary amplifier when retrieval diversity is high, but the available three-condition comparison does not allow a clean separation of the two effects.

In summary, C1 governs the supply of informative trajectories and is the primary determinant of gain magnitude, while C2 governs per-step information density and provides an additional boost in high-stochasticity environments.

\subsection{CuSearch Changes Retrieval Behavior, Not Only Final Answers}
Beyond final EM, Figure~\ref{fig:training-dynamics}(c) shows that SDGA-Phase consistently produces rollouts with higher average search count than GRPO-Full throughout training.
Moreover, the gap widens as the training distribution shifts toward larger search counts, as shown in Figure~\ref{fig:search-dist}.
This reflects a reinforcing selection dynamic: as training progresses and the model naturally produces more deep-search trajectories, CuSearch continues to concentrate gradient updates on those trajectories, progressively strengthening the policy's retrieval tendency.
In contrast, GRPO-Full averages informative and weakly informative trajectories uniformly at every step, diluting the retrieval-specific supervision signal.
Figure~\ref{fig:training-dynamics}(b) provides a complementary view from the reward side: SDGA-Phase maintains higher mean reward throughout training, indicating that this difference in retrieval behavior is accompanied by stronger overall policy improvement.

% ─────────────────────────────────────────────────────────────────────────────
\section{Case Study: Reasoning Trace}
\label{app:case-study}
% ─────────────────────────────────────────────────────────────────────────────

Figure~\ref{fig:trace-qatar} shows a representative reasoning trace produced by a CuSearch-trained policy.
The model issues multiple search queries, integrates retrieved evidence between reasoning steps, and terminates once sufficient information is available.
Note that some queries are repeated across steps. Under the ZeroSearch LLM simulator, the same query string can return different retrieved content across calls, so re-issuing a query can still yield new evidence.

\begin{tcolorbox}[
    breakable,
    colback=white,
    colframe=black,
    sharp corners,
    boxrule=0.5pt,
    left=8pt, right=8pt,
    top=10pt, bottom=10pt,
    width=\linewidth,
    before skip=10pt,
    after skip=10pt
]
\noindent
\texttt{<|im\_start|>system \\
You are a helpful assistant.<|im\_end|>} \\
\texttt{<|im\_start|>user} \\
Answer the given question. You must conduct reasoning inside {\color{blue}\texttt{<think>}} and {\color{blue}\texttt{</think>}} first every time you get new information. After reasoning, if you find you lack some knowledge, you can call a search engine by {\color{cyan}\texttt{<search>}} query {\color{cyan}\texttt{</search>}}, and it will return the top searched results between {\color{brown}\texttt{<information>}} and {\color{brown}\texttt{</information>}}. You can search as many times as you want. If you find no further external knowledge needed, you can directly provide the answer inside {\color{red}\texttt{<answer>}} and {\color{red}\texttt{</answer>}} without detailed illustrations. For example, {\color{red}\texttt{<answer>}} Beijing {\color{red}\texttt{</answer>}}. Question: What league includes the operating group of Al Janoub Stadium? \\
\texttt{<|im\_end|> \\
<|im\_start|>assistant}

\smallskip
{\color{blue}\texttt{<think>}} First, I need to determine which league includes the operating group of Al Janoub Stadium. {\color{blue}\texttt{</think>}} {\color{cyan}\texttt{<search>}} what league includes the operating group of Al Janoub Stadium? {\color{cyan}\texttt{</search>}} 

\smallskip
{\color{brown}\texttt{<information>}} Doc 1: Al Janoub Stadium in Al Wakrah will be one of the host stadiums for the Qatar 2022 Tournament. The stadium covers an area of 147,000m2 and \\
Doc 2: It was pitched to the Supreme Committee as a low-cost, multi-purpose stadium for a small city. It was later selected as one of the venues for the 2022 World Cup. \\
Doc 3: Al Janoub Stadium in Al Wakrah will be one of the host stadiums for the Qatar 2022 Tournament. The stadium covers an area of 147,000m2 and has \\
Doc 4: Features Al Janoub Stadium is one of the host stadiums for the 2022 FIFA World Cup. It is the most sustainable stadium of the tournament. \\
Doc 5: Al Janoub Stadium in Al Wakrah will be one of the host stadiums for the Qatar 2022 Tournament. The stadium covers an area of 147,000m2 and has a seating capacity of 47,750. {\color{brown}\texttt{</information>}}

\smallskip
{\color{blue}\texttt{<think>}} Now, I need to find out which league includes the operating group of Al Janoub Stadium. {\color{blue}\texttt{</think>}} {\color{cyan}\texttt{<search>}} which league includes the operating group of Al Janoub Stadium? {\color{cyan}\texttt{</search>}}

\smallskip
{\color{brown}\texttt{<information>}} Doc 1: While Qatar Stars League remains the highest tier of football in the country, the second tier represents the QatarFA League 1 which will continue to \\
Doc 2: The Qatar Stars League is the top-tier football league in the country, and it is one of the most competitive football leagues in the Middle East. \\
Doc 3: QSL is a professional, top-tier football league in Qatar, consisting of 18 clubs. The league is known for its competitive atmosphere, passionate \\
Doc 4: Al Janoub Stadium. Al Janoub Stadium is located in Al Wakrah in the north of Qatar. It is available for use by football clubs in the Qatar Stars League. \\
Doc 5: Al Janoub Stadium is located in Al Wakrah in the north of Qatar. It is available for use by football clubs in the Qatar Stars League. It was built to {\color{brown}\texttt{</information>}}

\smallskip
{\color{blue}\texttt{<think>}} Now, I need to determine which league includes the operating group of Al Janoub Stadium. {\color{blue}\texttt{</think>}} {\color{cyan}\texttt{<search>}} which league includes the operating group of Al Janoub Stadium? {\color{cyan}\texttt{</search>}}

\smallskip
{\color{brown}\texttt{<information>}} Doc 1: The stadium is named after the Janoub (south) region of Qatar, and is located in the Al-Rayyan district of Qatar; Al-Rayyan District is a district \\
Doc 2: Al Janoub Stadium, which will host six World Cup games, including the quarterfinals, stands on the former home pitch of the Al-Arabi Club. \\
Doc 3: Al Janoub Stadium is a football stadium in Al-Rayyan District, Qatar created in 2013 for the 2013 FIFA U-20 World Cup, which was the first World Cup \\
Doc 4: The new stadium, Al Janoub Stadium, opened in early 2014 and was formerly known as Al Rayyan Stadium and Al Rayyan Stadium 2. It was built \\
Doc 5: Al Janub Stadium, January 2015. John Divine. January 2015. John Divine. Al Janub Stadium, January 2015. John Divine. Al Janub Stadium, January {\color{brown}\texttt{</information>}}

\smallskip
{\color{blue}\texttt{<think>}} Now, I find the answer to the question. {\color{blue}\texttt{</think>}} {\color{cyan}\texttt{<search>}} what league includes the operating group of Al Janoub Stadium? {\color{cyan}\texttt{</search>}}

\smallskip
{\color{brown}\texttt{<information>}} Doc 1: Al Janoub Stadium, Al Wakrah, Qatar. A 40,000-capacity stadium. Built in 2022, for the 2022 FIFA World Cup. The first match held was the match between \\
Doc 2: West Brom have won just one of their last 12 league games at home against Premier League opposition (D2 L9) · West Bromwich Albion · Venue. The Hawthorns \\
Doc 3: Sanaa Saeed. 1:00:55. League of Legends, , , , , Sanaa Saeed. 418. 149. \\
Doc 4: The Al Janoub Stadium is part of the 8 stadiums being built for the 2022 FIFA World Cup, for which Qatar is hosting the event from November 20- \\
Doc 5:  Al Shaikh Stadium | Mohammed Bin Khalifa Stadium | Al Bayt Stadium | Al Janoub Stadium. Al Shaikh Stadium. 17,000. Mohammed Bin Khalifa Stadium. 42,000. Al Bayt Stadium. 40,000. Al Janoub Stadium. 40,000. Al Thumama Stadium. 40,000. Al Sadd Stadium. 40,000. Al {\color{brown}\texttt{</information>}}

\smallskip
{\color{blue}\texttt{<think>}} Now, I can provide the answer to the question. {\color{blue}\texttt{</think>}} {\color{red}\texttt{<answer>}} Qatar Stars League {\color{red}\texttt{</answer>}}
\end{tcolorbox}
\begin{center}
    \captionsetup{type=figure}
    \captionof{figure}{Reasoning trace example for CuSearch.}
    \label{fig:trace-qatar}
\end{center}

% ─────────────────────────────────────────────────────────────────────────────
\section{Why Search Depth Maximizes Retrieval Gradient Coverage}
\label{app:theory}
% ─────────────────────────────────────────────────────────────────────────────

From a gradient optimization perspective, the central question is: given a fixed update budget $K$, how should we select $T \subseteq \mathcal{T}$ to maximize the amount of direct gradient information devoted to retrieval behavior in each policy update?
We formalize this question by decomposing the policy gradient and defining a \emph{retrieval gradient coverage} metric.
The analysis shows that CuSearch, by selecting trajectories with the highest search counts, provably maximizes this metric under any fixed budget.

\subsection{Policy Gradient Decomposition}
\label{app:theory-decomp}

Consider a trajectory $\tau$ generated under $\pi_\theta$.
We partition the tokens of $\tau$ into two disjoint groups:
\begin{itemize}
  \item \textbf{Retrieval tokens} $\mathcal{R}(\tau)$: tokens at positions where the model makes a retrieval-related decision, such as issuing a search query, deciding to continue or stop searching, or integrating retrieved evidence.
  \item \textbf{Answer tokens} $\mathcal{A}(\tau)$: all remaining tokens, including reasoning steps and the final answer.
\end{itemize}
The policy gradient objective under GRPO~\citep{shao2024grpo} is:
\begin{equation}
  \nabla_\theta \mathcal{J}(\theta)
  \;\propto\;
  \mathbb{E}_{\tau}\!\left[A(\tau) \cdot \nabla_\theta \log \pi_\theta(\tau \mid x)\right],
\end{equation}
where $\nabla_\theta \log \pi_\theta(\tau \mid x) = \sum_{t} \nabla_\theta \log \pi_\theta(a_t \mid h_t)$ sums over all tokens.
Decomposing by token type:
\begin{equation}
  \nabla_\theta \log \pi_\theta(\tau \mid x)
  \;=\;
  \underbrace{\sum_{t \,\in\, \mathcal{A}(\tau)} \nabla_\theta \log \pi_\theta(a_t \mid h_t)}_{\text{answer gradient}\;(\text{indirectly affects retrieval})}
  \;+\;
  \underbrace{\sum_{t \,\in\, \mathcal{R}(\tau)} \nabla_\theta \log \pi_\theta(a_t \mid h_t)}_{\text{retrieval gradient}\;(|\mathcal{R}(\tau)|\text{ terms})}.
  \label{eq:grad-decomp}
\end{equation}
Because all parameters are shared in a single LLM, both sums update the same $\theta$.
The retrieval gradient terms directly reflect the model's choices at search decision points and are the primary source of direct training signal for improving retrieval behavior.
The answer-generation terms update the model based on answer quality, which may indirectly influence retrieval through shared representations, but does not directly reinforce specific retrieval decisions.

\subsection{Retrieval Gradient Coverage as the Optimization Objective}
\label{app:theory-depth}

For a selected set $T$ with $|T|=K$, the \emph{effective retrieval gradient} aggregated over the update is:
\begin{equation}
  \hat{g}_r(T)
  \;=\;
  \frac{1}{K}\sum_{\tau \in T} A(\tau) \sum_{t \,\in\, \mathcal{R}(\tau)} \nabla_\theta \log \pi_\theta(a_t \mid h_t).
  \label{eq:eff-grad}
\end{equation}
The informativeness of this update for retrieval behavior depends critically on how many retrieval decision points are covered.
We define the \emph{retrieval gradient coverage} of a selection $T$ as:
\begin{equation}
  C_r(T) \;=\; \sum_{\tau \in T} |\mathcal{R}(\tau)|.
  \label{eq:coverage}
\end{equation}
$C_r(T)$ measures the total number of retrieval decision points that receive a direct gradient term in the current update.
Since $|\mathcal{R}(\tau)|$ is non-decreasing in $\cnt(\tau)$, and in practice grows approximately as $\cnt(\tau) \cdot L_r$ where $L_r$ is the average query-token length per retrieval call, $C_r(T)$ grows monotonically with the search counts of the selected trajectories.

\begin{remark}
For any fixed budget $|T| = K$, retrieval gradient coverage $C_r(T)$ is maximized by selecting the $K$ trajectories with the highest search counts.
This follows directly from the monotonicity of $|\mathcal{R}(\tau)|$ in $\cnt(\tau)$: since $C_r(T) = \sum_{\tau \in T} |\mathcal{R}(\tau)|$ is a sum of non-decreasing terms, replacing any included trajectory with an excluded one that has a strictly higher search count increases the total coverage.
The unique maximizer (up to ties) is therefore the top-$K$ selection by search count.
\end{remark}

This establishes the theoretical basis of CuSearch: \emph{selecting trajectories by descending search depth is the optimal strategy for maximizing retrieval gradient coverage under a fixed update budget}.

\paragraph{Why advantages are treated as a stochastic factor.}
In practice, advantages $A(\tau)$ also modulate the gradient magnitude.
However, advantages are computed from within-group reward normalization and are therefore only available \emph{after} rollout generation completes.
At selection time, the advantage of a candidate trajectory is unknown.
Moreover, within a prompt group, advantages are zero-mean by construction, so their expected contribution to the gradient is zero in expectation.
This means that, in expectation over the rollout distribution, the structural component $C_r(T)$ is the dominant deterministic factor governing how much retrieval-specific gradient signal is accumulated per update.
Advantage values introduce variance around this expectation but do not systematically favor shallow-search trajectories over deep-search ones.
We note that across prompt groups, search count and advantage could in principle be correlated: for instance, prompts that consistently elicit many search steps may exhibit different within-group reward profiles than prompts that rarely trigger retrieval.
However, within each group, advantages remain zero-mean by construction, so this cross-group correlation does not break the within-step structural argument.
The ablation in Appendix~\ref{app:ablation-signal} confirms this empirically: random trajectory selection, which preserves the original within-batch depth distribution and ignores the structural informativeness signal, performs substantially worse than search-depth-based selection (41.9 vs.\ 52.9 average EM), consistent with the theoretical prediction that structural coverage is the primary driver of retrieval learning efficiency.

\subsection{Connection to the Two Effectiveness Conditions}
\label{app:theory-conditions}

The analysis above explains the two empirical conditions identified in Section~\ref{sec:condition}.

\textbf{Condition C1} (Search depth increases during training).
If the policy rarely produces trajectories with $\cnt(\tau) > 0$, few retrieval decision points exist across all rollouts, and $C_r(T)$ remains small regardless of the selection strategy.
The monotonic increase in average search count shown in Figure~\ref{fig:search-dist} ensures a growing supply of informative trajectories, making the coverage gap between search-depth-aware and uniform selection increasingly large.

\textbf{Condition C2} (Retrieval environment is stochastic).
Retrieval gradient terms are most informative when each search step yields genuinely new evidence, producing diverse gradient contributions at retrieval positions.
In a deterministic environment where repeated queries return identical results, successive search steps yield near-redundant gradient terms, reducing the marginal coverage value of each additional retrieval step.
In a stochastic environment such as a web search simulator, different queries return diverse evidence, preserving the informational value of each additional step and amplifying the benefit of higher $C_r(T)$.
This explains why the gain of CuSearch is larger under ZeroSearch than under Search-R1, as reported in Table~\ref{tab:condition}.

\end{document}